\documentclass[conference]{IEEEtran}
\IEEEoverridecommandlockouts
\usepackage{multirow}
\usepackage{hyperref}
\usepackage{cite}
\usepackage{amsmath,amssymb,amsfonts}
\usepackage{algorithmic}
\usepackage{graphicx}
\usepackage{textcomp}
\usepackage{makecell}
\usepackage{xcolor}
\usepackage{booktabs}
\usepackage{pifont}
\usepackage{caption}
\def\BibTeX{{\rm B\kern-.05em{\sc i\kern-.025em b}\kern-.08em
    T\kern-.1667em\lower.7ex\hbox{E}\kern-.125emX}}
\begin{document}

\title{ECViT: Efficient Convolutional Vision Transformer with Local-Attention and Multi-scale Stages\\}

\author{\IEEEauthorblockN{1\textsuperscript{st}  Zhoujie Qian}
\IEEEauthorblockA{\textit{School of Computer Engineering and Science} \\
\textit{Shanghai University}\\
Shanghai, China \\
qzj040331@shu.edu.cn}
}

\maketitle

\begin{abstract}
 Vision Transformers (ViTs) have revolutionized computer vision by leveraging self-attention to model long-range dependencies. However, ViTs face challenges such as high computational costs due to the quadratic scaling of self-attention and the requirement of a large amount of training data. To address these limitations, we propose the Efficient Convolutional Vision Transformer (ECViT), a hybrid architecture that effectively combines the strengths of CNNs and Transformers. ECViT introduces inductive biases such as  locality and translation invariance, inherent to Convolutional Neural Networks (CNNs) into the Transformer framework by extracting patches from low-level features and enhancing the encoder with convolutional operations. Additionally, it incorporates local-attention and a pyramid structure to enable efficient multi-scale feature extraction and representation. Experimental results demonstrate that ECViT achieves an optimal balance between performance and efficiency, outperforming state-of-the-art models on various image classification tasks while maintaining low computational and storage requirements. ECViT offers an ideal solution for applications that prioritize high efficiency without compromising performance.
\end{abstract}


\section{Introduction}
Transformers use self-attention\cite{b1} to model long-range dependencies, revolutionizing how models handle sequential data. The Vision Transformer (ViT)\cite{b2} treats images as sequences of patches and uses the self-attention to capture global dependencies, which has made a successful transition from natural language processing (NLP) to computer vision (CV). ViT and its variants have demonstrated state-of-the-art performance for various tasks including image classification, detection, and segmentation\cite{b3,b4,b5,b23,b24,b25,b26,b27,b28,b21,b29,b31,b34}.

Although ViT shows promising results, it faces challenges that drive further research. One major issue is its lack of certain desirable biases such as translation invariance and locality, which are inherent in Convolutional Neural Networks (CNNs)\cite{b6}. These biases allow CNNs to perform effectively on smaller datasets by leveraging local receptive fields, weight sharing, and spatial subsampling. In contrast, ViT’s absence of such biases results in underperformance when applied to smaller datasets. Another challenge is its high computational cost. ViT's self-attention mechanism involves pairwise comparisons of all image patches, which scales quadratically with the number of patches. This results in significant computational overhead, making it resource-intensive, especially for high-resolution images and large models.

\begin{figure}[tbp]
\centerline{\includegraphics[width=0.5\textwidth]{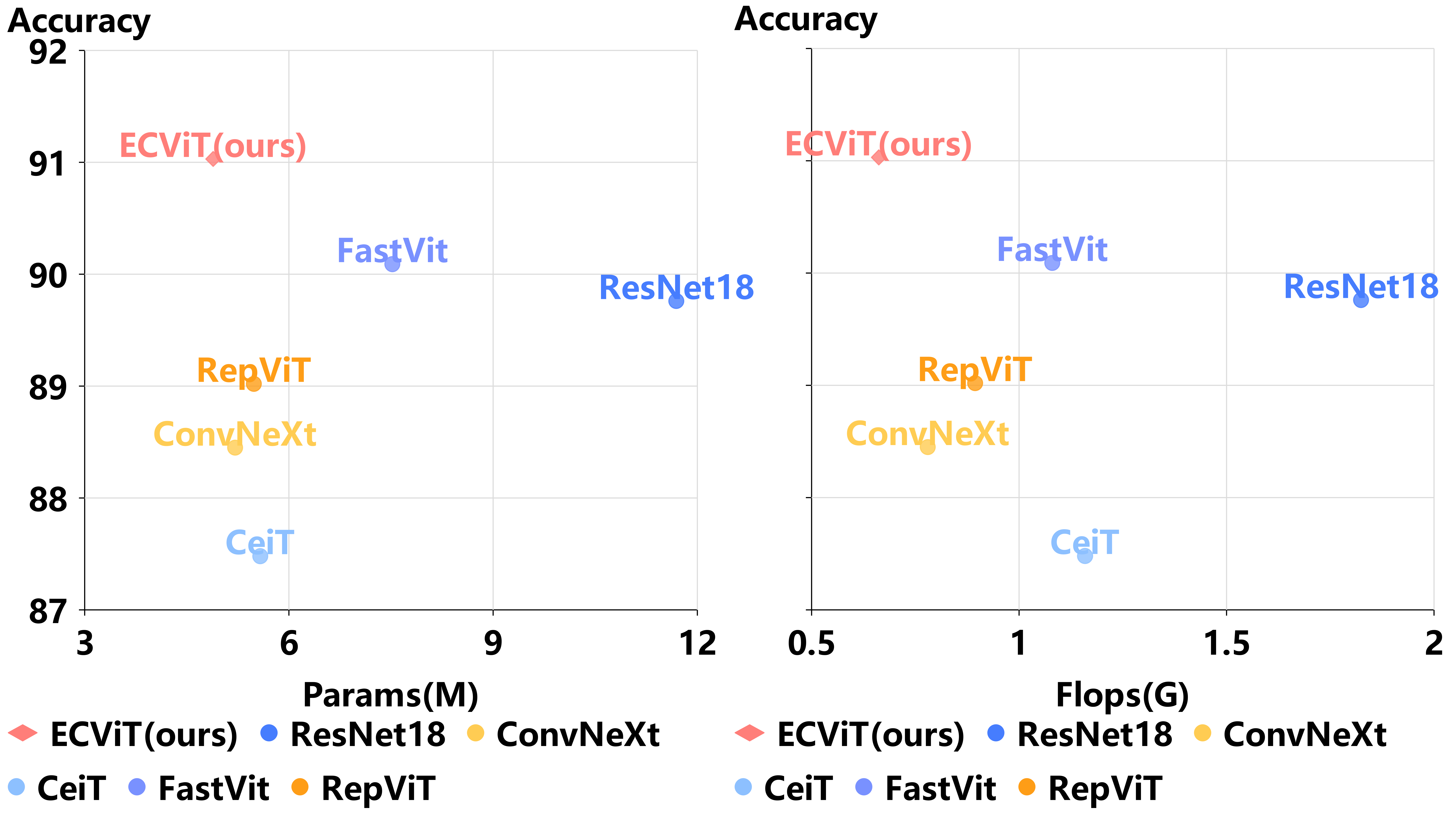}}
\caption{Performance comparison with the lightweight state-of-the-art models on CIFAR10. Our ECViT achieves the best balance between performance and efficiency.}
\label{result}
\end{figure}

To address these challenges, researchers have explored several enhancements. One promising approach is the integration of hybrid models that combine the strengths of both ViT and CNNs. Hybrid models aim to leverage the global modeling capabilities of self-attention mechanisms in ViTs while retaining the inductive biases of CNNs\cite{b7,b8,b26,b29,b30}. Another line of research focuses on improving the efficiency of the self-attention mechanism itself\cite{b9,b11,b31,b32,b33}. Techniques such as sparse attention\cite{b9,b11} and low-rank approximations\cite{b10} have been proposed to reduce the quadratic scaling of self-attention. 

For example, hybrid models like Convolutional Vision Transformers (CvT)\cite{b7} and LeViT\cite{b8} integrate convolutions with transformers to improve performance on smaller datasets and reduce computational complexity, bridging the gap between CNNs and pure transformer-based models. In another line of work, Linformer\cite{b9} approximates the self-attention matrix with a low-rank factorization, significantly reducing computational requirements. Similarly, Performer\cite{b10} introduces kernel-based methods to make self-attention computations linear in complexity. Additionally, Swin Transformer\cite{b11} tackles computational inefficiency by using a hierarchical structure and shifted windows. While these advancements have addressed some limitations of Vision Transformers, they still leave challenges that require further exploration. One significant issue is the trade-off between computational efficiency and the ability to capture both local and global features effectively. Another is that  these models typically modify specific modules of the ViT, such as tokenization or self-attention, rather than addressing the overall architecture in a more holistic manner.

To remedy this, we propose the Efficient Convolutional Vision Transformer (ECViT), which effectively combines the strengths of CNNs and transformers. To introduce certain inductive biases of CNNs, such as locality and translation invariance, to the ViT, our ECViT extracts patches from generated low-level features and enhances the transformer encoder with convolutional operations. Besides, we incorporate a pyramid structure\cite{b21,b22,b11,b31} within the transformer framework to generate multi-scale feature maps, further improving its ability to capture a diverse range of spatial information. Experimental results demonstrate that ECViT strikes the optimal balance between performance and efficiency, optimizing computational resources while maintaining high accuracy across various classification tasks.

In summary, our contributions are as follows: 
\begin{itemize}
    \item We design the Efficient Convolutional Vision Transformer (ECViT), which effectively combines the strengths of CNNs and transformers, achieving an optimal balance between performance and efficiency.
\end{itemize}

 \begin{itemize}
     \item We incorporate the desirable properties inherent in CNNs into the core operations of ViT. Additionally, our ECViT leverages local-attention and a pyramid structure to ensure more efficient feature extraction and representation. This hybrid architecture effectively captures both local patterns and long-range dependencies while maintaining computational efficiency.
 \end{itemize}

 \begin{itemize}
     \item Compared to previous state-of-the-art models, experimental results on various image classification datasets demonstrate the effectiveness of ECViT. It achieves outstanding performance while maintaining low storage and computational requirements, highlighting its efficiency and capability in resource-constrained environments.
 \end{itemize}
 

\section{Related Work}

\subsection{Convolutional Neural Networks}
Convolutional Neural Networks (CNNs) have been a cornerstone in the field of computer vision, excelling in visual recognition tasks. Initially introduced by LeCun et al. for handwritten digit recognition\cite{b12}, CNNs leverage convolutional kernels with shared weights, which provide translation invariance and efficient feature extraction. The advent of GPUs paved the way for deeper CNN architectures, with AlexNet\cite{b13} marking a breakthrough in large-scale image classification. Since then, the complexity of these networks has increased, incorporating innovations such as GoogLeNet's\cite{b14} Inception module, ResNeXt's multipath convolutions\cite{b16} and DenseNet's\cite{b17} dense connections, all of which have significantly improved performance. ResNet’s\cite{b15} skip connections were particularly influential, mitigating the vanishing gradient problem and enabling the training of very deep networks. Advances in convolutional layers, such as depthwise and deformable convolutions\cite{b18}, have further optimized CNNs for mobile devices and enhanced their flexibility in feature extraction. 
Our work seeks to leverage the strengths of CNNs and advanced convolutional operations to introduce beneficial inductive biases into ViT, thereby enhancing its accuracy and efficiency.

\begin{figure*}[tbp]
\centerline{\includegraphics[width=\textwidth]{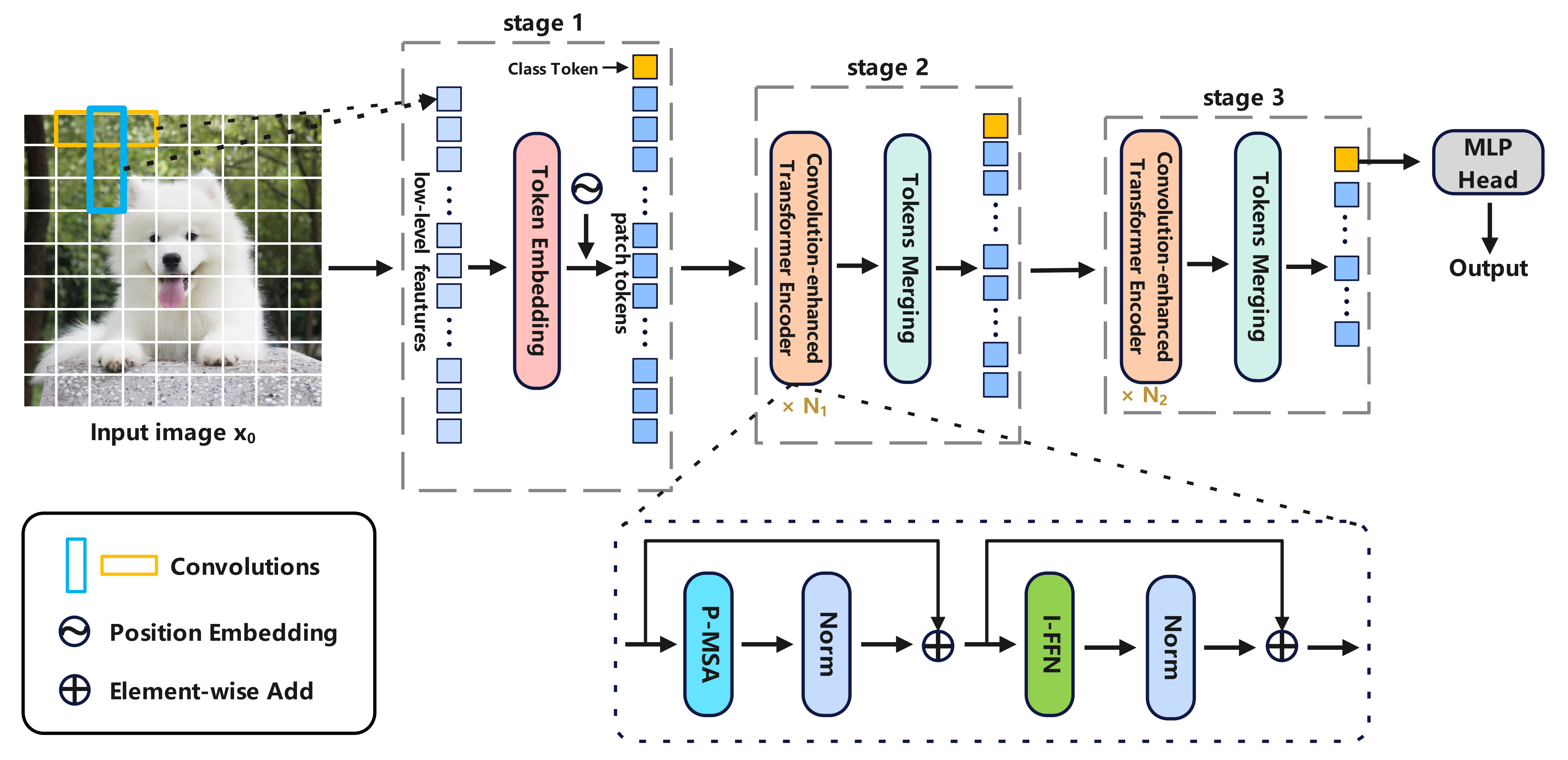}}
\caption{\textbf{Model overview.} Our model operates in three stages, each generating feature maps at different scales. The first stage employs a convolutional network to extract low-dimensional features, which are then transformed into a sequence of tokens, including a class token. The next two stages share a similar architecture, consisting of several convolution-enhanced transformer encoder layers and a merging layer. Each encoder contains two sub-layers: partitioned multi-head self-attention (P-MSA) and Interactive Feed-forward Network (I-FFN), with residual connections applied after each module. A merging layer is then applied to reduce the sequence length and increase the feature dimensionality. Finally, the class token is utilized for prediction through an MLP Head.}
\label{overview}
\end{figure*}

\subsection{Vision Transformers}
The Vision Transformer (ViT)\cite{b2} introduced by Dosovitskiy et al. in 2020 has gained considerable attention in the field of computer vision. ViT is a departure from traditional convolutional neural networks (CNNs), as it uses a pure transformer architecture, which was initially designed for natural language processing tasks, to handle image classification. By dividing an image into fixed-size patches and processing these patches as a sequence, ViT leverages the self-attention mechanism to capture long-range dependencies in images. The success of ViT demonstrated the potential of transformer-based models in vision tasks, which had been previously dominated by CNNs. Following ViT, several works have explored ways to improve its efficiency and performance. DeiT\cite{b3} addressed the data inefficiency issue by introducing a distillation technique, enabling ViT to perform well even with smaller datasets. Swin Transformer\cite{b11} further improved ViT by introducing a hierarchical design and local window-based self-attention, reducing the computational complexity and making the model more efficient for dense prediction tasks. In our research, we propose improvements to the original ViT structure, striving to balance the trade-off between its efficiency and accuracy.

\subsection{Vision Transformers with Convolution}
Recent advancements in Vision Transformers (ViTs) have explored various methods to enhance their performance by integrating convolutional operations. These approaches aim to improve local feature extraction while preserving the global context that transformers are known for.
The Convolutional Vision Transformer (CvT)\cite{b7} combines convolutional layers with transformers to improve local feature extraction while maintaining global context modeling. It uses convolutional tokens as input to better capture low-level features before applying global self-attention.
The Co-Scale Conv-Attentional Image Transformer (CaiT)\cite{b19} introduces "co-scaling," using convolutions at multiple scales to efficiently capture both local details and long-range dependencies.
The T2T-ViT\cite{b20} uses convolutions to preprocess the image into tokens, reducing token count and improving efficiency while preserving spatial hierarchies.

Based on these works, we propose to introduce convolutions to two primary parts of the ViT: (1) We utilize two convolutional layers to capture low-level features from the images and encode them into a sequence of tokens. (2) We substitute the feed-forward network with a convolutional network to strengthen the relationships between neighboring tokens in spatial dimension.

\subsection{Pyramid Structure}
The pyramid structure is widely used in computer vision to model multi-scale features, enhancing the representation of both global and local information. This design reflects the hierarchical nature of vision tasks, where features at different scales contribute to understanding complex patterns. 
In Vision Transformers, pyramid structures address the limitations of flat tokenization in models like ViT, where all tokens are treated equally regardless of spatial scale. Hierarchical representations in pyramid-based transformers capture both high-resolution details and contextual information more effectively.

For example, the Pyramid Vision Transformer (PVT)\cite{b21} reduces token resolution while increasing feature dimensions, improving efficiency in tasks like object detection and segmentation. Similarly, the Swin Transformer\cite{b11} uses a shifted window approach in a pyramid framework to improve computational efficiency and capture long-range dependencies. The Segformer\cite{b22} combines a pyramid transformer encoder with a lightweight decoder for semantic segmentation, achieving state-of-the-art performance with high efficiency. In our work, we progressively adjust the number of tokens and their feature dimensions to integrate the pyramid structure into ViT, optimizing resource usage without compromising performance.

\section{Method}
The overall architecture of our Efficient Convolutional Vision Transformer (ECViT) is depicted in Fig. \ref{overview}. The primary objective is to incorporate desirable inductive biases into the Transformer framework while integrating local-attention mechanism and multi-scale feature extraction. First, a convolutional network is utilized to extract low-dimensional features from input images and embed them as tokens. Next, these patch tokens are processed through multiple Transformer encoders, which consist of Partitioned Multi-head Self-Attention (P-MSA) and Interactive Feed-forward Network (I-FFN). Finally, to introduce a pyramid structure, a merging layer is employed to reduce the sequence length while increasing the feature dimensionality. 


\subsection{Image Tokenization with Low-level Feature Extraction}
The standard Vision Transformer splits an image into patches and directly tokenize them, which may result in a loss of low-level features such as texture and edges. To solve this limitation, our modified Image Tokenization generates tokens from feature maps instead of raw input images.

As shown in Fig. \ref{overview}, the Image Tokenization is comprised of two convolution layers and a max-pooling layer. A BatchNorm layer and activation function are attached to each convolution layer to enhance the expressive ability of features.

Our ECViT first extracts features from the image horizontally while downsampling, resulting in an output tensor \(f_h(x)\in \mathbb{R}^{ \frac{D_0}{2}\times\frac{H}{S} \times W  }\), where \( H \) is the height, \( W \) is the width, \( S \) is the stride, and \( D_0 \) is the number of enriched channels. The function to extract horizontal features can be denoted as: 

\begin{equation}
f_h(x) = GeLu(BN(Conv2d(x)))
\end{equation}

Next, the function \(f_v\) further extracts features in the vertical direction and reduces the width, resulting in an output tensor \(x_l \in \mathbb{R}^{ D_0\times \frac{H}{S} \times \frac{W}{S} }\). Following this, a Max-pooling is utilized to further extract important features and reduce data dimensions. The procedures can be noted as:
\begin{equation}
x_l = MaxPool(f_v(f_h(x)))
\end{equation}

Similar to the ViT, the learned low-level feature maps \(x_l\) are reshaped into a sequence of patches. Specifically, the total number of patches is \( N = \frac{H}{S} \times \frac{W}{S} \) in the spatial dimension. A positional embedding \( pos \in \mathbb{R}^{ N \times D } \) is then added to each element of the projected sequence, where \( N \) is the number of patches and \( D \) is the dimension of each patch. This positional embedding encodes spatial information and breaks the permutation invariance of the sequence. 
\begin{equation}
x' = Linear(Reshape(x_l))+pos
\end{equation}

Finally, a learnable class token is appended to the sequence of projected image patches to serve as a representation for classification tasks.


\begin{figure}[tbp]
\centerline{\includegraphics[width=0.5\textwidth]{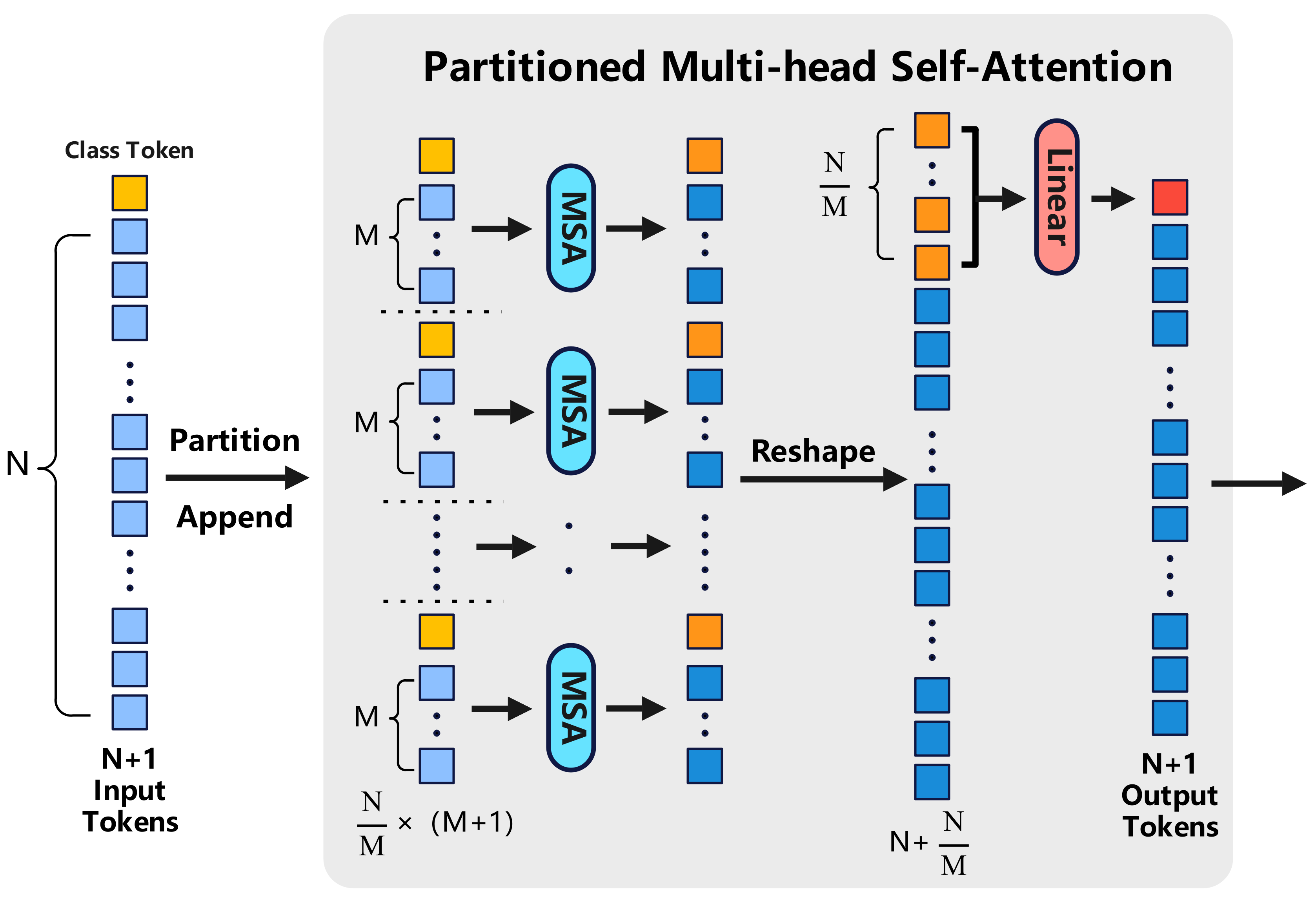}}
\caption{Partitioned Multi-head Self-Attention}
\label{MSA}
\end{figure}

\subsection{Partitioned Multi-head Self-Attention}
The standard Transformer architecture and its adaptation for image classification both perform global self-attention, which calculates the interactions between a token and all other tokens. This global computation results in quadratic complexity in relation to the number of tokens, which makes it impractical for many vision tasks that demand a large number of tokens for dense predictions or the representation of high-resolution images. 

To address this issue, we introduce the Partitioned Multi-Head Self-Attention (P-MSA), which performs self-attention within local blocks. As illustrated in Fig. \ref{MSA}, the patch tokens \( x_p^l \) are evenly partitioned into B non-overlapping sections where \( B = \frac{N}{M} \), with \( N \) being the total number of patch tokens and \(M\) the size of each block. Additionally, the class token \( x_c^l \) is appended to each block to facilitate the extraction of local features. The procedures can be summarized as follows: 
\begin{equation}
 x_c^l,x_p^l = Split(x_t^l)
\end{equation}
\begin{equation}
X_p = Partition(x_p^l)=[x_p^1,...x_p^b,...x_p^B]
\end{equation}
\begin{equation}
x_t^b = Concat( x_c^l,x_p^b)
\end{equation}
\begin{equation}
\hat{x_t^b} = MSA( x_t^b)
\end{equation}
\begin{equation}
\hat{X_p} = [\hat{x_t^1},...\hat{x_t^b},...\hat{x_t^B}]
\end{equation}

Following the attention computation, the individual class tokens from each block, which can be denoted as \([x_c^1,...x_c^b,...x_c^B]\), are merged into a single class token that effectively integrates the local information from all blocks, thereby improving the overall representation of the image. 
This approach reduces the computational burden while preserving the ability to capture both local and global context.

\begin{figure}[tbp]
\centerline{\includegraphics[width=0.5\textwidth,height=0.345\textwidth]{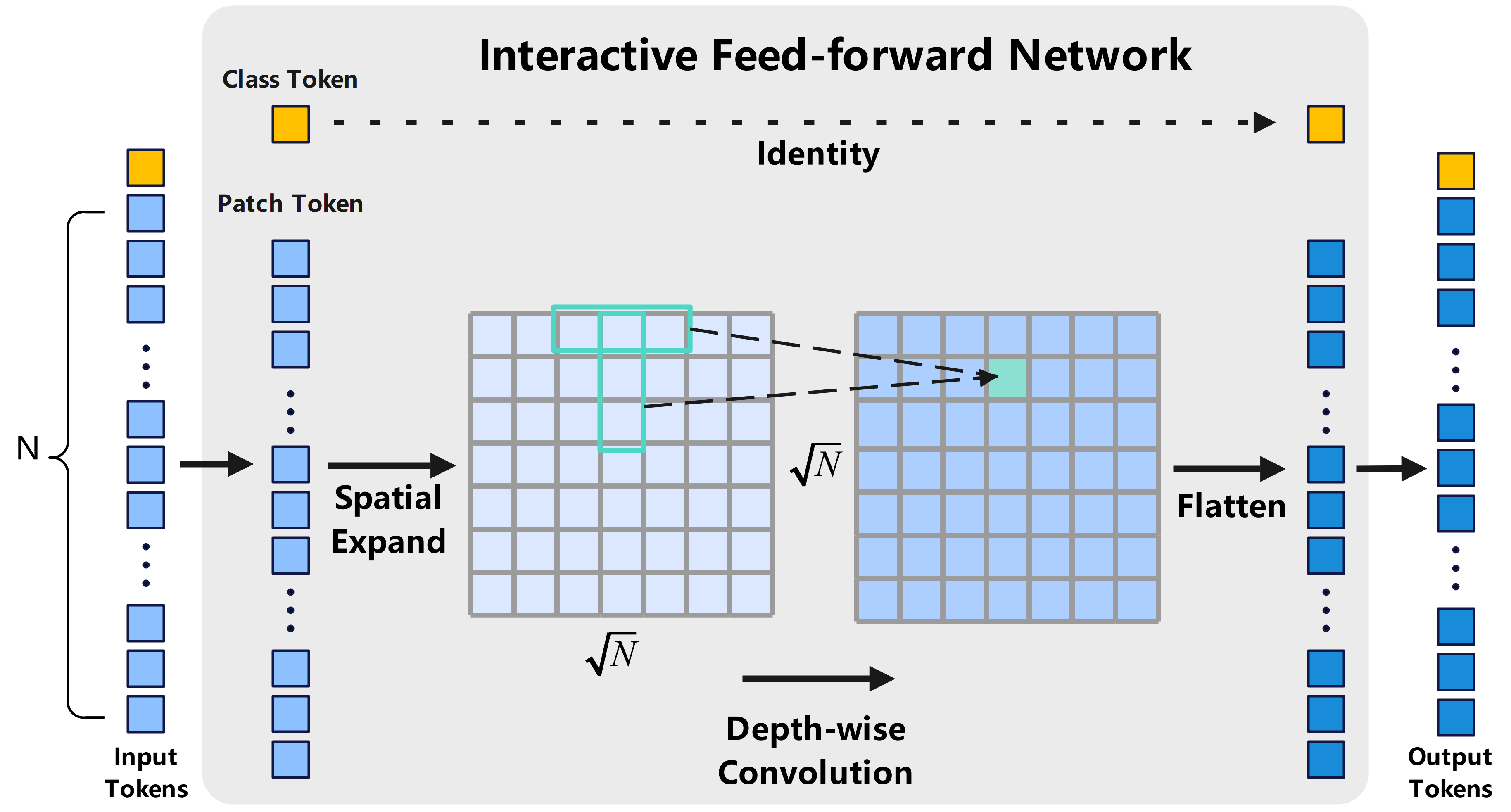}}
\caption{Interactive Feed-forward Network}
\label{FFN}
\end{figure}
\subsection{Interactive Feed-forward Network}
To leverage the strengths of CNNs in extracting local information and the Transformer’s capability to capture long-range dependencies and facilitate interaction between different non-overlapping blocks, we propose the Interactive Feed-Forward Network (I-FFN) layer. As illustrated in Fig. \ref{FFN}, the I-FFN module follows these steps, which can be noted as:
\begin{equation}
 x_c^l,x_p^l = Split(x_t^l)
\end{equation}
\begin{equation}
x_p^e = ExpandTo2d(x_p^l)
\end{equation}
\begin{equation}
x_p^c = GELU(BN(DepthConv(x_p^e)))
\end{equation}
\begin{equation}
x_p^f = FlattenTo1d(x_p^c)
\end{equation}
\begin{equation}
x_t^{l+1} = Concat(cls,x_p^f)
\end{equation}

 First, given the tokens \( x_t^l \in \mathbb{R}^{(N+1) \times D} \) generated by the preceding P-MSA module, we separate them into patch tokens \( x_p^l \in \mathbb{R}^{N \times D} \) and a class token \( x_c^l \in \mathbb{R}^D \). Next, the patch tokens are expanded to \( x_p^e \in \mathbb{R}^{\sqrt{N} \times\sqrt{N} \times D} \) along the spatial dimensions, based on their relative positions in the original image.

Then, two depth-wise separable convolutions, with kernel sizes of \((k, 1)\) and \((1, k)\), are applied to the rearranged patch tokens. This step enhances the local features and facilitates the integration of information across different spatial dimensions.

Finally, the patch tokens are flattened into a sequence \( x_p^f \in \mathbb{R}^{N \times D} \) and concatenated with the class token, resulting in \( x_t^{l+1} \in \mathbb{R}^{(N+1) \times D} \), preserving the original shape.

\subsection{Tokens Merging}
As mentioned in the overview, a merging layer is utilized after the last transformer encoder of each stage to modify the token feature dimensions and the number of tokens, thereby incorporating the pyramid structure into the Transformer framework.

Through Tokens Merging, we progressively reduce the sequence length while enhancing the feature dimensionality, allowing tokens to capture increasingly complex visual patterns across expanding spatial areas.These procedures can be noted as:

\begin{equation}
 x_c^l,x_p^l = Split(x_t^l)
\end{equation}
\begin{equation}
\hat{x_p^l} = MaxPool(x_p^l)
\end{equation}
\begin{equation}
\hat{x_t^l} = Concat(cls,\hat{x_p^l})
\end{equation}
\begin{equation}
x_t^{l+1} = Linear(\hat{x_t^l})
\end{equation}

After a series of transformer encoders, the tokens \( x_t^l \in \mathbb{R}^{(N+1) \times D} \) capture both global and local features. We begin by separating them into patch tokens \( x_p^l \in \mathbb{R}^{N \times D} \) and a class token \( x_c^l \in \mathbb{R}^D \).

Next, we employ a MaxPooling layer to further process the patch tokens \( x_p^l \) into \( \hat{x_p^l} \in \mathbb{R}^{\frac{N}{K} \times D} \), where \( K \) is the kernel-size of MaxPooling. This operation serves to down-sample the spatial dimensions while preserving the most significant features by selecting the maximum value within a specified window. This pooling step helps in maintaining robustness to spatial variations and consolidates the feature representation by focusing on dominant patterns.

Then, the patch tokens and the class token are reunited into \( \hat{x_t^l} \in \mathbb{R}^{(\frac{N}{K}+1) \times D} \), preserving the global context provided by the class token.

At last, a linear layer is applied to expand the feature dimension of the downsampled tokens, transforming \( \hat{x_t^l} \) into a higher-dimensional space \( x_t^{l+1} \in \mathbb{R}^{(\frac{N}{K}+1) \times D'} \). This transformation increases the representational capacity of the tokens, enabling them to capture more intricate patterns and relationships within the image.

\section{Experiments}
In this section, we assess the performance of the ECViT on some middle-scale image classification datasets and evaluate its transferability to various downstream tasks. We minimize the storage and computational requirements of the model, ensuring that it remains efficient and effective in scenarios with extremely limited computational resources, such as edge devices or environments with low processing power. Furthermore, we conduct comprehensive ablation studies to verify the effectiveness of the proposed architectural design.
\subsection{Experimental Settings}
\paragraph{\textbf{Network Architecture}}
After meticulously designing the number of Transformer blocks and the hidden feature dimension at each stage, we introduce the lightweight ECViT. The details of our ECViT are as follows: 

\begin{enumerate}
    \item The Image Tokenization utilizes two depth-wise separable convolutions, with kernel sizes of 7\(\times\)1 and 1\(\times\)7 and a stride of 2. The number of output channels is set to 32. The max pooling operation uses a kernel size of 3\(\times\)3 and a stride of 2.
    \item Every stage consists of 8 convolution-enhanced encoder blocks, each meticulously optimized for efficiency. In the P-MSA, the partition size is set to 7. Within the I-FFN, a pair of 3\(\times\)1 and 1\(\times\)3 depth-wise separable convolutions with a stride of 1 is applied to refine spatial feature representations. Additionally, a max pooling layer with a kernel size of 4\(\times\)4 and a stride of 4 is used to further reduce spatial dimensions and computational complexity. The hidden feature dimensions are progressively increased, with stage 2 set to 224 and stage 3 to 256, ensuring an optimal balance between model capacity and efficiency.
    \item The lightweight design of ECViT allows it to achieve excellent performance with minimal computational overhead. With only 4.888M parameters and 0.698G Flops, ECViT is highly suitable for deployment in resource-constrained environments, such as mobile devices or edge computing scenarios, without compromising on accuracy or effectiveness.
\end{enumerate}
\begin{table*}[htbp]
\caption{Performance comparison with the lightweight state-of-the-art models on various classification datasets. The first two highest accuracies are bolded. Our ECViT achieves the best balance between performance and efficiency.}
\centering
\resizebox{\textwidth}{!}{
\begin{tabular}{c|c|cc|cccccccccc} 
\toprule
Group & Model & \makecell{Params\\(M)} & \makecell{Flops\\(G)}  & \makecell{CIFAR\\10} & \makecell{CIFAR\\100} & SVHN & \makecell{Fashion\\MNIST} & Food & GTSRB & \makecell{FGVC\\Aircraft} & Imagenette  &\\ 
\midrule
\multirow{3}{*}{CNNs} 
& ResNet18\cite{b23} & 11.690 & 1.824 & 89.76 & 66.57 & 92.84 & 93.69 & 68.94 & 94.56 & 54.57 & 87.84 & \\ 
& EfficientNetV2-B0\cite{b24} & 7.024 & 0.649 & \textbf{91.99} & \textbf{71.67} & \textbf{97.11} & \textbf{94.27} & \textbf{72.04} & \textbf{97.60} & \textbf{62.52} & \textbf{88.20} & \\ 
& ConvNeXt-V2\cite{b25} & 5.210 & 0.780 & 88.45 & 66.66 & 95.74 & 93.27 & 64.21 & 96.14 & 33.75 & 82.80 & \\ 
\midrule
\multirow{6}{*}{Transformers}
& ViT-T/16\cite{b2} & 5.679 & 1.079 & 76.35 & 49.96 & 48.05 & 88.70 & 47.78 & 65.24 & 23.76 & 76.53 & \\ 
& PVTv2-B0\cite{b21} & 3.662 & 0.529 & 90.44 & 67.87 & 95.99 & 93.36 & 66.44 & 95.83 & 60.51 & 86.80 & \\ 
& CeiT-T\cite{b26} & 5.579 & 1.159 & 87.48 & 69.19 & 96.41 & 93.63 & 71.39 & 96.82 & 61.54 & 86.82 & \\
& FastViT-t8\cite{b27} & 7.520 & 1.080 & 90.09 & 67.87 & 95.78 & \textbf{94.38} & \textbf{72.01} & 92.07 & 45.75 & 87.03 & \\ 
& RepViT-m1\cite{b28} & 5.485 & 0.894 & 89.02 & 67.65 & 96.20 & 93.44 & 67.94 & 96.14 & 54.75 & 85.95 & \\
& \textbf{Ours: ECViT} & 4.888 & 0.698 & \textbf{91.03} & \textbf{70.88} & \textbf{97.96} & 93.84 & 69.39 & \textbf{96.93} & \textbf{62.28} & \textbf{88.73} & \\ 
\bottomrule
\end{tabular}
}
\label{results}
\end{table*}

\paragraph{\textbf{Training}}
All of our experiments are conducted on the NVIDIA Tesla P100 GPUs. AdamW optimizer is used with the weight decay of 0.05 for our ECViT. We train our model with an initial learning rate of 0.01 and a total batch size of 64 for 30 epochs, with a cosine learning rate decay scheduler.

\paragraph{\textbf{Datasets}}
We test our ECViT on various datasets including superordinate level classification (CIFAR10, CIFAR100, Imagenette) and Primary recognition task (SVHN, FashionMNIST), fine-grained recognition  (Food, GTSRB, FGVCAircraft). The CIFAR10 dataset is composed of 10 classes, containing 50,000 training samples and 10,000 validation samples. The CIFAR100 dataset, with 100 classes, also includes 50,000 training samples and 10,000 validation samples. The SVHN (Street View House Numbers) dataset, which contains 10 classes, has 73,257 training samples and 26,032 validation samples. The FashionMNIST dataset, including 10 classes, comprises 60,000 training samples and 10,000 validation samples. The Food dataset, which includes 101 classes, contains 75,000 training samples and 25,000 validation samples. The GTSRB (German Traffic Sign Recognition Benchmark) dataset, including 43 classes, contains a total of 39,209 training samples and 12,630 validation samples. The FGVCAircraft  dataset is composed of 30 classes, containing 6,667 training samples and 3,333 validation samples. Last, the Imagenette dataset, a curated subset of the ImageNet dataset, consists of 10 classes and contains 9,469 training samples and 3,925 validation samples. These datasets provide a diverse range of classes and sample sizes for training and validating our model in various computer vision tasks.

\subsection{Results on Image Classification}
For comparison, we select a diverse set of models, including Convolutional Neural Networks (CNNs) such as ResNets\cite{b23}, EfficientNets\cite{b24}  and ConvNeXt-V2\cite{b25}, Transformers like ViT\cite{b2} and PvT\cite{b21}, as well as hybrid models such as CeiT\cite{b26}, FastViT\cite{b27} and RepViT\cite{b28}. These models represent a range of architectures, from traditional convolution-based approaches to more recent transformer-based and hybrid designs, allowing us to comprehensively evaluate the performance of our proposed ECViT. The results of these models on the aforementioned test datasets are presented in Table \ref{results}. 

\paragraph{\textbf{ECViT vs convolution-based models}}
With significantly fewer parameters (4.888M) and a low FLOPs count (0.698G) compared to convolution-based models such as ResNet18 (11.690M, 1.824G), ConvNeXt (5.210M, 0.780G), and EfficientNetV2 (7.024M, 0.649G), ECViT stands out as a lightweight and efficient alternative, making it ideal for resource-constrained environments. On CIFAR10, ECViT achieves 91.03\% accuracy, outperforming ResNet18 (89.76\%) and ConvNeXt (88.45\%), while remaining competitive with EfficientNetV2 (91.99\%). On SVHN, ECViT achieves the highest accuracy (97.96\%), surpassing all benchmark models. 

ECViT combines the merits of the ViTs and CNNs, showing strong generalization across datasets. It consistently delivers high accuracy with minimal computational overhead, making it a strong candidate for tasks requiring a trade-off between performance and efficiency. While EfficientNetV2 achieves slightly higher accuracy in some datasets, it does so at the cost of higher parameter counts. Conversely, ECViT strikes a better balance, particularly on datasets like SVHN, FashionMNIST, GTSRB FGVCAircraft and Imagenette,  where it outperforms or matches competitive models while being lighter and faster.

\paragraph{\textbf{ECViT vs Transformers}}
Compared to ViT and PvT, our ECViT demonstrates a superior balance between computational efficiency and performance across various datasets. ECViT is significantly more efficient than ViT-tiny (5.679M, 1.079G) while delivering much higher accuracy. For example, on CIFAR10, ECViT achieves 91.03\% accuracy, far surpassing ViT-tiny (76.35\%), and on CIFAR100, ECViT achieves 70.88\%, significantly outperforming ViT-tiny (49.96\%). Similarly, compared to PvT-v2, which has slightly fewer parameters (3.662M) and lower FLOPs (0.529G), ECViT achieves superior accuracy across datasets, including SVHN (97.96\% vs. 95.99\%), GTSRB(96.93\% vs. 95.83\%), FGVCAircraft(62.28\% vs. 60.51\%) and Imagenette(88.73\% vs. 86.80\%).

When compared to Ceit, FastViT and RepViT, which incorporate various optimizations for efficiency and performance, our ECViT stands out with its ability to maintain a higher level of accuracy without sacrificing computational cost. On CIFAR10, ECViT achieves 91.03\% accuracy, outperforming CeiT-T (87.48\%), FastViT-t8 (90.09\%) and RepViT-m1(89.02\%). Similarly, on CIFAR100, ECViT delivers 70.88\% accuracy, surpassing CeiT-T (69.19\%), FastViT-t8 (67.87\%) and RepViT-m1(67.65\%). 

ViT frameworks often struggle with smaller datasets due to their reliance on large-scale pretraining. In contrast, ECViT excels on smaller datasets such as CIFAR10, CIFAR100, GTSRB, FGVCAircraft and Imagenette, without any pretraining. ECViT outperforms other state-of-the-art ViT models on most datasets while maintaining a more lightweight architecture and lower computational demands. This is due to its hybrid design, which incorporates convolutional inductive biases into the transformer architecture. As a result, ECViT offers an ideal solution for applications that require high efficiency without compromising performance.

\subsection{Ablation Study}
To further investigate the effectiveness of the proposed modules, we conduct ablation studies on the main components of Image Tokenization, Partitioned Multi-head Self-Attention, Interactive Feed-forward Network and Tokens Merging. All of our ablation experiments are based on the CIFAR10 dataset.
\paragraph{\textbf{Different types of Tokenization module}}
The factors influencing the Tokenization module include the convolution kernel size, the presence of Max-pooling, BatchNorm layers, and activation functions. In Table \ref{Tokenization}, we evaluate different convolution configurations and select the first as the optimal choice. The results, shown in Table \ref{Token}, demonstrate that Max-pooling, BatchNorm layers, and the GELU activation significantly enhance performance during training. Consequently, we adopt the optimal structure (detailed in the first row) for all subsequent experiments.

\begin{table}[tbp]
\caption{Ablation study results on the proposed modules}
\begin{center}
\begin{tabular}{cccc|c}
\toprule
\textbf{Tokenization} & \textbf{P-MSA} & \textbf{I-FFN}& \textbf{Tokens Merging}&\textbf{accuracy} \\
\midrule
\ding{55} & \ding{55} &\ding{55}& \ding{55} & 76.35 \\
\ding{51} & \ding{55} &\ding{55}& \ding{55} &81.96 (\textcolor{red}{+5.61}) \\
\ding{51} & \ding{51} &\ding{55}& \ding{55} & 82.80 (\textcolor{red}{+6.45}) \\
\ding{51} & \ding{51} &\ding{51}& \ding{55} & 89.78 (\textcolor{red}{+13.43}) \\
\ding{51} & \ding{51} &\ding{51}& \ding{51} & 91.03 (\textcolor{red}{+14.68}) \\
\bottomrule
\end{tabular}
\label{module}
\end{center}
\end{table}

\begin{table}[tbp]
\caption{Ablation study results on the type of Convolution}
\begin{center}
\begin{tabular}{cc|c  c|c}
\toprule
\textbf{kernel} & \textbf{stride} & \makecell{\textbf{Params}\\\textbf{(M)}} & \makecell{\textbf{Flops}\\\textbf{(G)}} & \textbf{accuracy} \\
\midrule
(7,1) + (1,7) & (2,1)+(1,2) & 4.888 & 0.698 &  91.03\\
(7,7) & (2,2) & 4.906 & 0.703 & 90.19 (\textcolor{green}{-0.84}) \\
(5,5) & (2,2) & 4.904 & 0.673 & 89.75 (\textcolor{green}{-1.28}) \\
\bottomrule
\end{tabular}
\label{Tokenization}
\end{center}
\end{table}

\begin{table}[tbp]
\caption{Ablation study results on the type of Tokenization}
\begin{center}
\begin{tabular}{ccc|c}
\toprule
\textbf{maxpool} & \textbf{BN} & \makecell{\textbf{Activation }\\\textbf{Functions}} & \textbf{accuracy} \\
\midrule
\ding{51} & \ding{51} & GELU & 91.03 \\
\ding{51} & \ding{51} & RELU & 90.21 (\textcolor{green}{-0.82}) \\
\ding{51} & \ding{55} & GELU & 89.35 (\textcolor{green}{-1.68}) \\
\ding{51} & \ding{51} & \ding{55} & 88.79 (\textcolor{green}{-2.24}) \\
\ding{55} & \ding{51} & GELU & 87.23 (\textcolor{green}{-3.80}) \\
\bottomrule
\end{tabular}
\label{Token}
\end{center}
\end{table}

\paragraph{\textbf{Partitioned Multi-head Self-Attention}}
In the P-MSA module, the block size determines the number of tokens involved in the attention computation. We evaluate different block sizes and compare the performance with and without class token appending. The results are presented in Table \ref{PMSA}. Considering the trade-off between computational complexity and accuracy, we select a block size of 7. By appending the class token to each block, performance improves from 85.54\% to 91.03\%, demonstrating that enhancing the extraction of local features is highly beneficial.

\begin{table}[tbp]
\caption{Ablation study results on P-MSA}
\begin{center}
\begin{tabular}{c c| c| c}
\toprule
\textbf{block size} & \makecell{\textbf{Class}\\\textbf{Appending}} & \makecell{\textbf{Flops}\\\textbf{(G)}} & \textbf{accuracy} \\
\midrule
7 & \ding{51} & 0.698 & 91.03 \\
\ding{55} & \ding{55} & 1.047 (\textcolor{green}{+0.349}) & 91.25 (\textcolor{red}{+0.22}) \\
7 & \ding{55} & 0.634 (\textcolor{red}{-0.064}) &  85.54 (\textcolor{green}{-5.49}) \\
14 & \ding{51}  & 0.732 (\textcolor{green}{+0.034}) &  90.87 (\textcolor{red}{-0.16}) \\
28 & \ding{51}  & 0.762 (\textcolor{green}{+0.064}) &  91.22 (\textcolor{red}{+0.19}) \\
\bottomrule
\end{tabular}
\label{PMSA}
\end{center}
\end{table}

\paragraph{\textbf{Interactive Feed-forward Network} }
In the I-FFN module, the kernel size determines the region within which patch tokens establish local correlations. To evaluate its impact on performance, we experiment with different kernel sizes, as shown in Table \ref{IFFN}. Larger kernel sizes enable patch tokens to capture broader local correlations, potentially improving feature representation. However, excessively large kernels can introduce unnecessary complexity and increase computational overhead. Therefore, we select a kernel size of 3 and employ Factorized Convolutions to strike an optimal balance between computational complexity, parameter count, and accuracy.

\paragraph{\textbf{Tokens Merging}}
We compare the performance with and without the Token Merging module. By adopting the Token Merging module, the number of parameters is reduced by 27.74\%, while the performance improves from 89.78\% to 91.03\%. This demonstrates that reducing the sequence length while enhancing feature dimensionality contributes significantly to the final image representation.

\begin{table}[tbp]
\caption{Ablation study results on I-FFN}
\begin{center}
\begin{tabular}{c c | c}
\toprule
\textbf{kernel} & \textbf{BN} & \textbf{accuracy} \\
\midrule
(3,1) + (1,3) & \ding{51}& 91.03 \\
(3,1) + (1,3) & \ding{55}& 89.97 (\textcolor{green}{-1.06}) \\
(3,3) & \ding{51}&  90.05 (\textcolor{green}{-1.02})\\
(5,1) + (1,5) & \ding{51}&  91.25 (\textcolor{red}{+0.22})\\
(7,1) + (1,7) & \ding{51}&  91.38 (\textcolor{red}{+0.35})\\
\bottomrule
\end{tabular}
\label{IFFN}
\end{center}
\end{table}

\begin{table}[tbp]
\caption{Ablation study results on the depth}
\begin{center}
\begin{tabular}{c|c c| c}
\toprule
\textbf{depth}  & \makecell{\textbf{Params}\\\textbf{(M)}} & \makecell{\textbf{Flops}\\\textbf{(G)}} & \textbf{accuracy} \\
\midrule
8 & 4.888 & 0.698 & 91.03 \\
4 & 2.497 (\textcolor{red}{-2.391}) & 0.388 (\textcolor{red}{-0.310}) & 88.34 (\textcolor{green}{-2.69}) \\
16 & 9.672 (\textcolor{green}{+4.784}) & 1.319 (\textcolor{red}{+0.621}) & 91.53 (\textcolor{red}{+0.50}) \\
\bottomrule
\end{tabular}
\label{depth}
\end{center}
\end{table}

\section{Conclusion}
In this paper, we introduce ECViT, a lightweight vision Transformer that effectively integrates the strengths of CNNs with the capabilities of Transformers in modeling long-range dependencies. To further improve efficiency, we incorporate local-attention and a pyramid structure into ECViT, enabling more effective feature extraction and representation. ECViT achieves an optimal balance between performance and computational efficiency, without the need for pretraining.


\begin{thebibliography}{00}
\bibitem{b1} A. Vaswani, N. Shazeer, N. Parmar, J. Uszkoreit, L. Jones, A. N. Gomez, Ł. Kaiser, and I. Polosukhin, “Attention is all you need,” in \textit{Advances in Neural Information Processing Systems (NeurIPS)}, 2017, pp. 5998–6008.
\bibitem{b2} A. Dosovitskiy, L. Beyer, A. Kolesnikov, D. Weissenborn, X. Zhai, T. Unterthiner, M. Dehghani, et al., “An image is worth 16x16 words: Transformers for image recognition at scale,” in \textit{International Conference on Learning Representations (ICLR)}, 2021.
\bibitem{b3} H. Touvron, M. Cord, M. Douze, F. Massa, A. Sablayrolles, and H. Jégou, “Training data-efficient image transformers \& distillation through attention,” in \textit{International Conference on Machine Learning (ICML)}, 2021.
\bibitem{b4} N. Carion, F. Massa, G. Synnaeve, N. Usunier, A. Kirillov, and S. Zagoruyko, “End-to-end object detection with transformers,” in \textit{European Conference on Computer Vision (ECCV)}, 2020, pp. 213–229.
\bibitem{b5} S. Zheng, J. Lu, H. Zhao, X. Zhu, Z. Luo, Y. Wang, Y. Fu, J. Feng, T. Xiang, P. Torr, and L. Zhang, “Rethinking semantic segmentation from a sequence-to-sequence perspective with transformers,” in \textit{Proceedings of the IEEE/CVF Conference on Computer Vision and Pattern Recognition (CVPR)}, 2021, pp. 6881–6890.
\bibitem{b6} M. Raghu, T. Unterthiner, S. Kornblith, C. Zhang, and A. Dosovitskiy, “Do vision transformers see like convolutional neural networks?” in \textit{Advances in Neural Information Processing Systems (NeurIPS)}, 2021, vol. 34, pp. 12116–12128.
\bibitem{b7} Z. Wu, C. Xu, D. Zhang, W. Zhang, and X. Li, “CvT: Introducing convolutions to vision transformers,” in \textit{Proceedings of the IEEE/CVF International Conference on Computer Vision (ICCV)}, 2021, pp. 22–31.
\bibitem{b8} B. Yu, H. Fan, J. Guo, Z. Wang, and S. Wu, “LeViT: A vision transformer in ConvNet’s clothing for faster inference,” in \textit{Proceedings of the IEEE/CVF Conference on Computer Vision and Pattern Recognition (CVPR)}, 2021, pp. 12119–12129.
\bibitem{b9} S. Wang, J. Li, M. Chen, and J. Smola, “Linformer: Self-attention with linear complexity,” in \textit{Advances in Neural Information Processing Systems (NeurIPS)}, 2020, vol. 33, pp. 6298–6309.
\bibitem{b10} K. Choromanski, V. Likhosherstov, D. Dohan, X. Song, A. Gane, T. Sarl\'{o}s, P. Hawkins, J. Davis, A. Mohiuddin, L. Kaiser, and M. Belanger, “Rethinking attention with Performers,” in \textit{Proceedings of the International Conference on Learning Representations (ICLR)}, 2021.
\bibitem{b11} Z. Liu, H. Hu, Y. Lin, Z. Yao, Z. Xie, Y. Wei, J. Ning, Y. Cao, Z. Zhang, L. Dong, F. Wei, and B. Guo, “Swin transformer: Hierarchical vision transformer using shifted windows,” in \textit{Proceedings of the IEEE/CVF International Conference on Computer Vision (ICCV)}, 2021, pp. 10012–10022.
\bibitem{b12} Y. LeCun, B. Boser, J. Denker, D. Henderson, R. Howard, W. Hubbard, and L. Jackel, “Backpropagation applied to handwritten zip code recognition,” in \textit{Neural Computation}, vol. 1, no. 4, 1989, pp. 541–551.
\bibitem{b13} A. Krizhevsky, I. Sutskever, and G. E. Hinton, “ImageNet classification with deep convolutional neural networks,” in \textit{Advances in Neural Information Processing Systems (NIPS)}, 2012
\bibitem{b14} C. Szegedy, W. Liu, Y. Jia, P. Sermanet, S. Reed, D. Anguelov, D. Erhan, V. Vanhoucke, and A. Rabinovich, “Going deeper with convolutions,” in \textit{Proceedings of the IEEE Conference on Computer Vision and Pattern Recognition (CVPR)}, 2015, pp. 1–9.
\bibitem{b15} K. He, X. Zhang, S. Ren, and J. Sun, “Deep residual learning for image recognition,” in \textit{Proceedings of the IEEE/CVF Conference on Computer Vision and Pattern Recognition (CVPR)}, 2016
\bibitem{b16} S. Xie, R. Girshick, P. Dollár, Z. Tu, and K. He, “Aggregated residual transformations for deep neural networks,” in \textit{Proceedings of the IEEE/CVF Conference on Computer Vision and Pattern Recognition (CVPR)}, 2017
\bibitem{b17} G. Huang, Z. Liu, L. van der Maaten, and K. Q. Weinberger, “Densely connected convolutional networks,” in \textit{Proceedings of the IEEE/CVF Conference on Computer Vision and Pattern Recognition (CVPR)}, 2017
\bibitem{b18} J. Dai, Y. Li, K. He, and J. Sun, “Deformable convolutional networks,” in \textit{Proceedings of the IEEE/CVF International Conference on Computer Vision (ICCV)}, 2017
\bibitem{b19} Z. Touvron, M. Cord, M. Douze, and A. Joulin, “Going deeper with Image Transformers,” in \textit{Proceedings of the IEEE/CVF Conference on Computer Vision and Pattern Recognition (CVPR)}, 2021, pp. 12140–12149.
\bibitem{b20} Y. Chen, J. Li, H. Zhang, Y. Li, Y. Xie, and L. Wang, “Token-to-Token Vision Transformers,” in \textit{Proceedings of the IEEE/CVF International Conference on Computer Vision (ICCV)}, 2021, pp. 1034–1043.
\bibitem{b21} Z. Zhang, Y. Li, H. Zhang, Y. Li, Y. Xie, and L. Wang, “Pyramid Vision Transformer: A Versatile Backbone for Dense Prediction without Convolutions,” in \textit{Proceedings of the IEEE/CVF International Conference on Computer Vision (ICCV)}, 2021
\bibitem{b22} E. Xie, W. Wang, Z. Yu, A. Anandkumar, P. Dollár, and L. Yuan, “Segformer: Simple and efficient design for semantic segmentation with transformers,” in \textit{Advances in Neural Information Processing Systems (NeurIPS)}, 2021
\bibitem{b23} K. Hara, H. Kataoka, and Y. Satoh, “Learning spatio-temporal features with 3D residual networks for action recognition,” in \textit{Proceedings of the IEEE/CVF International Conference on Computer Vision (ICCV)}, 2017
\bibitem{b24} M. Tan and Q. V. Le, “EfficientNet: Rethinking model scaling for convolutional neural networks,” in \textit{Proceedings of the International Conference on Machine Learning (ICML)}, 2019
\bibitem{b25} Z. Wu, Y. Zhang, Y. Zhang, and Y. Wei, “ConvNeXt V2: Co-designing of self-supervised learning techniques and architectural improvements,” in \textit{Proceedings of the IEEE/CVF Conference on Computer Vision and Pattern Recognition (CVPR)}, 2024
\bibitem{b26} J. Wang, Y. Li, and Z. Zhang, “Incorporating Convolution Designs into Visual Transformers,” in \textit{Proceedings of the IEEE/CVF Conference on Computer Vision and Pattern Recognition (CVPR)}, 2021
\bibitem{b27} S. Li, Y. Zhang, and Z. Zhang, “FastViT: A hybrid vision architecture for fast and efficient inference,” in \textit{Proceedings of the IEEE/CVF International Conference on Computer Vision (ICCV)}, 2023
\bibitem{b28} A. Wang, H. Chen, Z. Lin, J. Han, and G. Ding, “RepViT: Revisiting Mobile CNN From ViT Perspective,” in \textit{Proceedings of the IEEE/CVF Conference on Computer Vision and Pattern Recognition (CVPR)}, 2024, pp. 15909–15920
\bibitem{b29} H. Fan, B. Xiong, C.-Y. Wu, S. Mangalam, Y. Li, Z. Yan, J. Malik, and C. Feichtenhofer, “Multiscale Vision Transformers,” in \textit{Proceedings of the IEEE/CVF International Conference on Computer Vision (ICCV)}, 2021, pp. 6824–6835.
\bibitem{b30} W. Jiang, Z. Hou, L. Cao, M. Zhang, H. Bao, and Y. Chen, “ConvFormer: Closing the Gap Between CNN and Vision Transformers,” in \textit{Proceedings of the IEEE/CVF Conference on Computer Vision and Pattern Recognition (CVPR)}, 2023, pp. 10310–10319.
\bibitem{b31} S. Peng, M. Huang, W. Guo, Z. Wang, and Y. Sun, “PiRaMA: Pyramid Range Attention Mechanism for Efficient Vision Transformer,” in \textit{Proceedings of the IEEE/CVF Conference on Computer Vision and Pattern Recognition (CVPR)}, 2023, pp. 12301–12310.
\bibitem{b32} Z. Nguyen, M. Mahadevan, and A. Smola, “Faster Attention Is All You Need,” in \textit{Proceedings of Advances in Neural Information Processing Systems (NeurIPS)}, 2022.
\bibitem{b33} A. Roy, M. Vaswani, and A. Roychowdhury, “Efficient Attention Mechanism for Vision Transformers Using Low-Rank Decomposition,” in \textit{Proceedings of the IEEE/CVF International Conference on Computer Vision (ICCV)}, 2023, pp. 12347–12356.
\bibitem{b34} A. Kumar, R. Singh, and K. Balasubramanian, “EfficientViT: Lightweight Vision Transformers for Resource-Constrained Devices,” in \textit{Proceedings of the IEEE/CVF International Conference on Computer Vision (ICCV)}, 2024, pp. 1234–1243.
\end{thebibliography}
\end{document}